\documentclass[table, dvipsnames]{article} 
\pdfoutput=1 
\usepackage{collas2024_conference,times}
\usepackage{subcaption}
\usepackage{booktabs}
\usepackage{multirow}
\usepackage{amsmath}
\usepackage{makecell}
\usepackage{listings}
\usepackage[frozencache,cachedir=.]{minted}
\usepackage{adjustbox}
\usepackage{caption}
\usepackage{comment}
\usepackage{colortbl}
\usepackage{xcolor}

\usepackage{amsmath,amsfonts,bm}

\def\eqref#1{equation~\ref{#1}}

\def\1{\bm{1}}

\DeclareMathAlphabet{\mathsfit}{\encodingdefault}{\sfdefault}{m}{sl}
\SetMathAlphabet{\mathsfit}{bold}{\encodingdefault}{\sfdefault}{bx}{n}

\usepackage{hyperref}
\hypersetup{
    colorlinks=true,
    linkcolor=red,
    filecolor=magenta,
    urlcolor=blue,
    citecolor=purple,
    pdftitle={Overleaf Example},
    pdfpagemode=FullScreen,
    }

\definecolor{bg}{rgb}{0.95,0.95,0.91}

\title{ARCLE: The Abstraction and Reasoning Corpus Learning Environment for Reinforcement Learning}

\newcommand\CoAuthorMark{\footnotemark[\arabic{footnote}]}

\author{Hosung Lee\textsuperscript{1}\thanks{Equal Contribution}, Sejin Kim\textsuperscript{1}\protect\CoAuthorMark, 
Seungpil Lee\textsuperscript{1}, Sanha Hwang\textsuperscript{1}, Jihwan Lee\textsuperscript{1}, Byung-Jun Lee\textsuperscript{2}\thanks{Corresponding to: byungjunlee@korea.ac.kr, sundong@gist.ac.kr}, Sundong Kim\textsuperscript{1}\CoAuthorMark \\ 
\textsuperscript{1}Gwangju Institute of Science and Technology\\
\textsuperscript{2}Korea University \\
}

\collasfinalcopy 
\newcommand{\arclerepo}{https://github.com/ConfeitoHS/arcle}

\begin{document}

\maketitle

\begin{abstract}

This paper introduces ARCLE, an environment designed to facilitate reinforcement learning research on the Abstraction and Reasoning Corpus (ARC). 
Addressing this inductive reasoning benchmark with reinforcement learning presents these challenges: a vast action space, a hard-to-reach goal, and a variety of tasks.
We demonstrate that an agent with proximal policy optimization can learn individual tasks through ARCLE. 
The adoption of non-factorial policies and auxiliary losses led to performance enhancements, effectively mitigating issues associated with action spaces and goal attainment. 
Based on these insights, we propose several research directions and motivations for using ARCLE, including MAML, GFlowNets, and World Models.

\end{abstract}

\section{Introduction}

\begin{figure}[htbp!]
    \centering 
    \includegraphics[width=0.93\columnwidth]{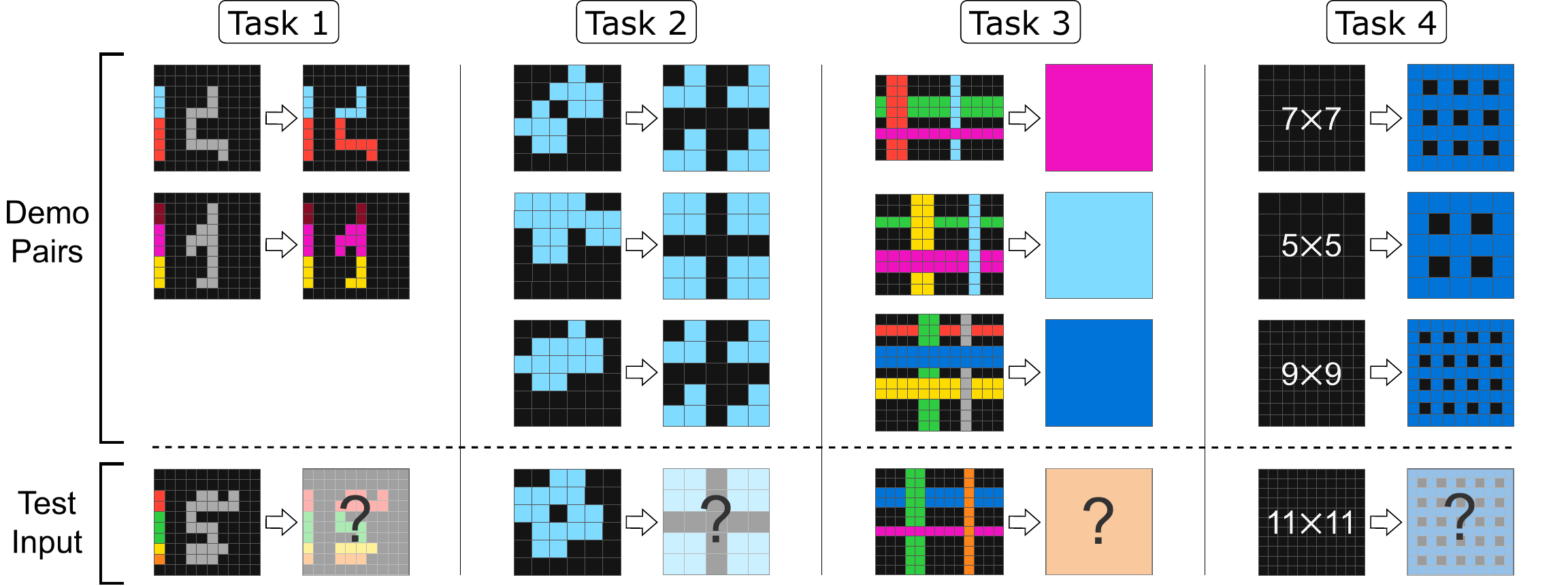} 
    \caption{Four different ARC tasks are presented, each requiring analysis through its provided demonstration pairs. The identified rule from this analysis must then be applied to a test input grid to produce the answer (test output) grid, which is currently blurred for demonstration purposes. The specific rules for each task are as follows: Task 1 modifies all gray grids within a row to match the color found in the far-left column of that same row. Task 2 relocates four identical cyan objects appropriately, each no larger than 2$\times$2 in size. Task 3 determines the color of the topmost line in a stack of overlapping horizontal and vertical lines, and outputs a single pixel of this color. Task 4 transforms the Test Input grid by coloring all but the pixels at the intersections of even-numbered rows and columns in blue.}
\label{fig:arc_example}
\end{figure}

We introduce ARCLE (ARC Learning Environment) as a reinforcement learning (RL) environment designed for the Abstraction and Reasoning Challenge (ARC) benchmark~\citep{chollet2019ARC}. This benchmark assesses agents' ability to infer rules from given grid pairs and predict the outcome for a test grid, as illustrated in Figure \ref{fig:arc_example}.
ARC is designed to test abstraction and reasoning skills, making it a touch benchmark within the domain. 
Despite various attempts to conquer ARC's complexities through program synthesis and reasoning using large language models, RL-based approaches are surprisingly rare (Section~\ref{subsec:solving_ARC}). 
We believe this scarcity is due to the lack of a dedicated RL environment tailored for ARC. 
To fill this gap, we created ARCLE based on Gymnasium~\citep{farama2023gymnasium} to tackle the benchmark.

From RL's standpoint, ARC is considered highly challenging. 
The typical difficulties include (1) a vast action space, (2) a hard-to-reach goal, and (3) a variety of tasks. 
While other RL benchmarks (e.g., robotics, financial trading, recommender systems, video games) might feature one of these challenges, ARC encompasses all, showing its difficulty. 
ARCLE is designed to help researchers navigate these challenges, offering a unique testbed for RL strategies. 

\paragraph{Vast action space}
ARC stands out with its vast action space by allowing a variety of actions such as coloring, moving, rotating, or flipping pixels. This feature creates a large set of possibilities, complicating the development of optimal strategies for RL agents. Such a vast action space demands innovative approaches to navigate effectively.

\paragraph{Hard-to-reach goal}
ARC tasks are uniquely challenging because success is measured by the ability to replicate complex grid patterns accurately, not by reaching a physical location or endpoint. This requires a deep understanding of the task rules and an ability to apply them precisely. Designing effective reward systems is particularly challenging in this context, as progress is not easily quantified. Each ARC task demands not just strategic action but also a nuanced comprehension of the underlying patterns and rules.

\paragraph{Variety of tasks}
ARC's wide array of tasks necessitates broad generalization, a stark contrast to benchmarks like Atari, which focus on mastering single games.\footnote{Atari benchmarks hosts 57 games, each with its goal. Solutions such as Rainbow DQN~\citep{mnih2013playing}, R2D2~\citep{revaud2019r2d2}, MuZero~\citep{schrittwieser2020mastering}, and Agent57~\citep{badia2020agent57} focus on mastering single games.} 
This diversity calls for adaptive and varied strategies, highlighting ARC's demand for advanced RL methods.

ARCLE is an environment that helps overcome the challenges of ARC and paves new pathways for AI research, seamlessly linking abstract reasoning in ARC with the adaptability in RL. Our initial experiments highlight the capability of RL to address specific tasks within ARC, indicating the potential necessity for advanced techniques such as meta-RL, generative models, or model-based RL algorithms. Thus, ARCLE stands out as a platform for testing RL solutions, prompting an in-depth exploration of the challenges ARC presents.
\section{Related Works}
\subsection{Solving ARC}
\label{subsec:solving_ARC}

Since the unveiling of the ARC~\citep{chollet2019ARC}, approaches ranging from the development of similar benchmarks~\citep{qi2021pqa, kim2022playgrounds, xu2023graphs} to domain-specific languages and program synthesis~\citep{banburski2020dreaming, acquaviva2022communicating, assouel2022object, alford2021neural, witt2023divide, ainooson2023approach} have been explored to extend its applicability and enhance learning strategies. 
These efforts have deepened our understanding of ARC’s challenges, highlighting the complexity of devising comprehensive solutions. 
The recent shift towards leveraging Large Language Models (LLMs), incorporating strategies from natural language processing to detailed task context integration~\citep{camposampiero2023abstract, xu2024llms, moskvichev2023conceptARC, mitchell2024comparing, lee2024reasoning}, underscores LLMs' potential in addressing ARC’s intricacies.

However, the performance of research on the ARC utilizing program synthesis and LLMs has not fully met expectations, often due to its logical flaw~\citep{lee2024reasoning}. This has prompted a pivot towards reinforcement learning as a novel approach, albeit its application to ARC has been limited. Notable attempts include using RL strategies in program synthesis~\citep{butt2024codeit} and exploring imitation learning~\citep{park2023unraveling}. 
The introduction of ARCLE opens up new possibilities for advancing research on the ARC using RL. 

\subsection{RL environments similar to ARCLE}

Among the myriad RL environments, those featuring a \textbf{vast action} space similar to ARCLE's are prominently observed in game-based settings, such as PySC2~\citep{vinyals2017starcraft}, where the diversity of actions, determined by mouse click locations, mirrors the flexible action format in ARC. 
Similarly, environments designed for recommendation systems (e.g., RecSim, RecoGym) and complex multi-step planning tasks (e.g., Super Mario Bros~\citep{kauten2018supermario}, NLE~\citep{kuttler2020nethack}) may not exhibit wide action spaces at each state but encapsulate the challenge of \textbf{hard-to-reach goal} through the necessity of sequential decision-making to achieve success. 
In parallel, the breadth of tasks within ARCLE resonates with the diverse objectives found in robotics environments like Meta-World~\citep{yu2020meta}, RLBench~\citep{james2020rlbench}, and CALVIN~\citep{mees2022calvin}, underscoring the complexity and \textbf{variety of tasks} that ARCLE introduces to RL research.
\section{ARCLE: ARC Learning Environment}

\begin{figure}[htbp!]
    \centering
    \includegraphics[width=0.8\columnwidth]{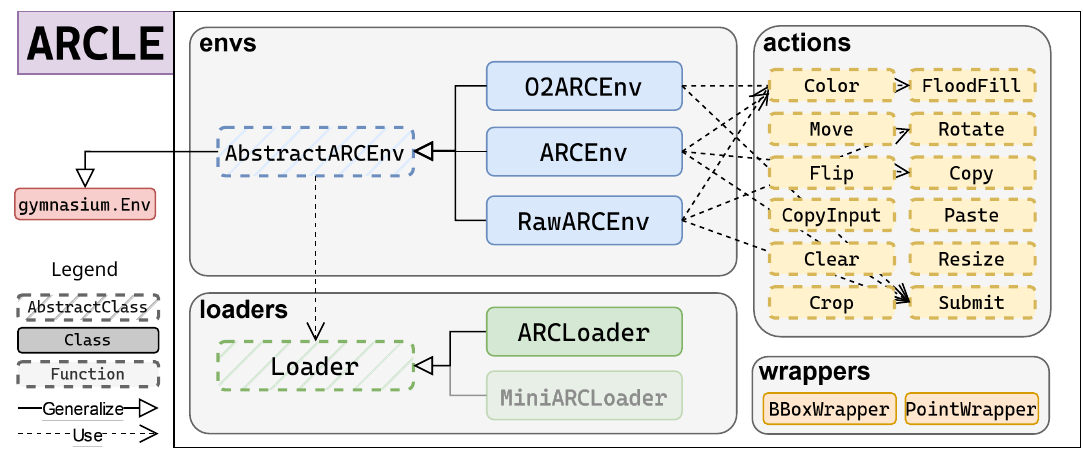}
    \caption{Framework of ARCLE. The package consists of components: envs, actions, loaders, and wrappers.}
    \label{fig:arcle_framework}
\end{figure}

ARCLE is a reinforcement learning (RL) environment package, implemented in Gymnasium~\citep{farama2023gymnasium}, designed for RL approaches on Abstraction and Reasoning Corpus (ARC). RL agents on the ARCLE environments learn to solve tasks by selecting actions to edit the grid (to be submitted) to the environment state. As Figure \ref{fig:arcle_framework} illustrates, ARCLE comprises three main components: \textbf{envs}, \textbf{loaders}, \textbf{actions}, and auxiliary \textbf{wrappers} which modify the environment's action or state space. 
The following explanation is based on the terms in Table \ref{tab:notations}.

The \textbf{envs} component consists of a base class of ARCLE environments, and its three derivatives. 
\texttt{AbstractARCEnv} inherits Gymnasium's \texttt{Env} class to provide reinforcement learning environment features and defines the ARC-specific structure of action and state space and user-definable methods. Its implementations, \texttt{O2ARCEnv}, \texttt{ARCEnv} and \texttt{RawARCEnv} provide embodied action and observation spaces. 
\texttt{O2ARCEnv} constructs the state and action space according to the O2ARC interface (See Appendix~\ref{subsec:o2arc}), and likewise, \texttt{ARCEnv} offers the testing web interface developed by \citet{chollet2019testinginterface}. 
\texttt{RawARCEnv} restricts the action space to color modifications or grid size changes, providing a more constrained and monotonic learning environment.

Next, the \textbf{loaders} component provides functionalities to supply the ARC dataset to ARCLE environments. This component comprises the base \texttt{Loader} class defining interface requirements to ARCLE environments and their implementations. \texttt{ARCLoader} feeds the ARC dataset to any ARCLE environment and defines how the ARC dataset should be parsed from files and how the parsed dataset should be picked. Likewise, to load a similar dataset to the ARC, one can inherit the \texttt{Loader} class and specify how to parse and sample. We provide the \texttt{MiniARCLoader} which loads Mini-ARC dataset~\citep{kim2022playgrounds} upon an ARCLE environment, as an example usage of \texttt{Loader} class. 

Last, \textbf{actions} component includes a variety of functions capable of changing environment state, called \textit{operation}. Each environment in ARCLE contains several operations to be used in an environment by agents on the environment. 
Since ARCLE currently implemented actions on the O2ARC interface, it contains more actions (e.g., \texttt{Move}, \texttt{Rotate}, \texttt{Flip}) than the original ARC testing interface~\citep{chollet2019testinginterface}. 

We focus on explaining \texttt{O2ARCEnv} in the following sections, which encompasses most operations by ARCLE.

\begin{figure}[htbp!]
    \centering
    \includegraphics[width=0.8\columnwidth]{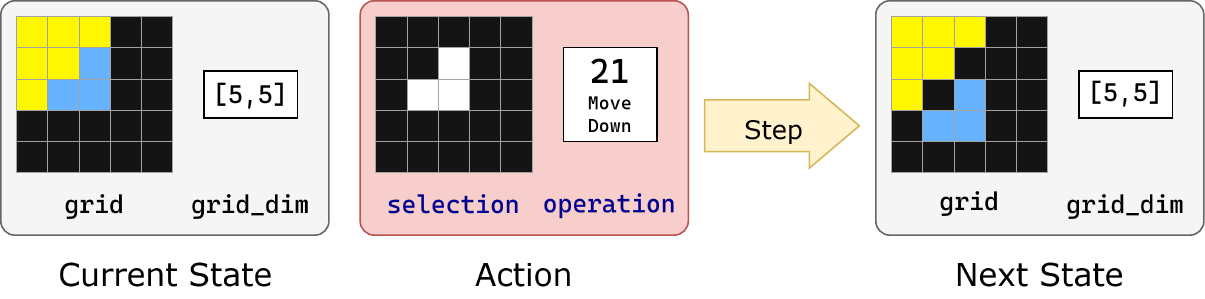}
    \caption{The state transition process of ARCLE.}
    \label{fig:arcle_state_transition}
\end{figure}

\begin{table*}[t!]
\centering
\caption{Variables in action and state spaces and their definition.}
\label{tab:notations}
\resizebox{0.85\linewidth}{!}{
\begin{tabular}{l|l|l} \toprule
    Variable Space & Name & Definition  \\ \midrule
    Action & \texttt{operation} & Integer index representing edit method of environment state (e.g., \texttt{grid}, \texttt{clip})\\
     & \texttt{selection} & Binary mask that specifies where a \texttt{operation} to be applied \\ \midrule
    State & \texttt{input} & Input grid of demonstration pair or test pair \\
    & \texttt{input\_dim} & Dimension (height, width) of \texttt{input} \\
    & \texttt{grid} & Editable output grid of demonstration pair or test pair \\  
     & \texttt{grid\_dim} & Dimension (height, width) of \texttt{grid} \\  
    & \texttt{clip} & Clipboard grid \\
    & \texttt{clip\_dim} & Dimension (height, width) of \texttt{clip}\\ \midrule
    State& \texttt{selected} & Binary array which represents currently selecting pixels for object-oriented operations \\
    (\texttt{object\_states})& \texttt{active} & Boolean variable of whether last operation was an object-oriented operation \\
    & \texttt{object} & Backed-up pixels of specified pixels of \texttt{grid} for object-oriented operations\\
    & \texttt{object\_sel} & Binary mask of exact shape which pixels of \texttt{object} that user has specified \\ 
    & \texttt{object\_dim} & Dimension (height, width) of bounding box of \texttt{object} and \texttt{object\_sel} \\ 
    & \texttt{object\_pos} & Left-top position of bounding box of \texttt{obejct} on the \texttt{grid}\\ 
    & \texttt{rotation\_parity} & Binary value for consistency over serial rotations \\ 
    & \texttt{background} & Pixels remaining in the \texttt{grid} excluding \\ \midrule 
    Answer & \texttt{answer} & Answer grid of test input grid \\ 
    (Hidden to agents) & \texttt{answer\_dim} & Dimension (height, width) of answer grid\\
    \bottomrule
\end{tabular}
}
\end{table*}

\subsection{Actions}
Actions in ARCLE are defined to enable editing of the output grid for a given task, consisting of \texttt{operation} and \texttt{selection}. 
\texttt{operation} represents an integer that specifies the method of editing (functions contained in the actions component in Figure \ref{fig:arcle_framework}), and \texttt{selection} is a binary mask that denotes the area of the grid affected by the edit.

By defining ARCLE's actions through \texttt{operation} and \texttt{selection} as illustrated by the action in the middle of Figure \ref{fig:arcle_state_transition}, we have standardized various types of actions within the same structure. 
Notably, the actions in ARCLE can affect a single pixel, contiguous multiple pixels, or even non-contiguous pixels, accommodating these possibilities through employing the binary mask selection. 
Furthermore, by separating \texttt{operation} and \texttt{selection}, it accommodates the possibility of determining selection conditioned by the chosen operation autoregressively.

\begin{figure}[htbp!]
    \centering
    \includegraphics[width=0.9\columnwidth]{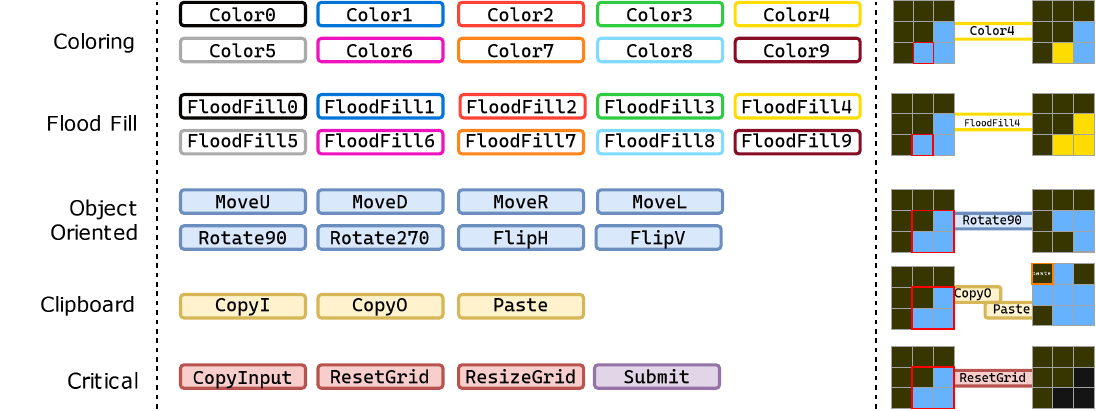}
    \caption{Every \texttt{operation} assigned in \texttt{O2ARCEnv} (version 0.2.5). The categories of operations (left), available operations (middle), and application examples of operations (right) are shown.}
    \label{fig:arcle_o2arc_ops}
\end{figure}

Currently, 35 operations are available in \texttt{O2ARCEnv} (Figure~\ref{fig:arcle_o2arc_ops}). 
When an agent specifies the operation index, the corresponding one is executed. 
Specifically, operation indices 0--9 represent to \texttt{Color} the selected pixels (by \texttt{selection}) with one color among the ten colors used in ARC, while 10--19 denote a \texttt{Flood Fill} based on Depth-First Search (DFS) in the selected pixels. 
Object-oriented operations not present in the original ARC testing interface~\citep{chollet2019testinginterface}, such as \texttt{Move}, \texttt{Rotate}, and \texttt{Flip}, are assigned to 20--23 (up, down, right, left), 24--25 (counterclockwise, clockwise), and 26--27 (horizontal, vertical), respectively. 
Additionally, 28--30 correspond to \texttt{Copy} and \texttt{Paste}, and 31--34 are assigned to operations that cause breaking changes in the states like duplicating the test input (\texttt{CopyInput}), clearing the grid (\texttt{ResetGrid}), changing the grid size (\texttt{ResizeGrid}), and submitting (\texttt{Submit}).

Subclassing the environments allows customizing available operations by adding or removing them in the same format. For a detailed description of every operation in ARCLE, see Appendix~\ref{subsec:action_description}.

\subsection{States \& Observations}
All environments included in ARCLE are designed with the assumption to be Markov Decision Processes (MDP). 
Therefore, every parameter used in changing the environment's state is given to agents in the environment, so observations and states can be considered equivalent. 
The basic state space of an environment within ARCLE consists of the \texttt{input} and \texttt{grid}. Input represents the test input grid of an ARC task, so it is fixed unless a new task is assigned to an environment. Grid is initially set as the test input grid of a task, and an agent edits this by selecting actions.

Depending on which operations an environment adopts, the state of the environment can be different. For instance, if an environment includes \texttt{Copy} operation, the environment should include additional variables of the copied part: \texttt{clip}. Hence in \texttt{O2ARCEnv}, more variables are included in the state, to support \texttt{Copy} and object-oriented operations such as \texttt{Move}. 
These object-oriented actions from the O2ARC interface are supplemented with \texttt{selected}, \texttt{object}, \texttt{object\_pos} and \texttt{background}. Descriptions of these variables are depicted in \ref{tab:notations}. While the agent performs object-oriented operations in a row, \texttt{object} and \texttt{background} works as two layers; \texttt{object} is overlayed on the \texttt{background} at \texttt{object\_pos}. For the detailed mechanism described in Section \ref{subsec:technical_details}.

\subsection{Rewards}
The built-in reward currently offered in ARCLE is the sparse reward. 
This reward grants $1$ when the agent performs the \texttt{submit} action and the state space's grid exactly matches the task's answer grid, and $0$ if even a single pixel differs. 
This sparse reward approach can hinder the learning of an agent whose total reward sum remains $0$ as there is a unique answer per task. 
To counteract this, an auxiliary reward was designed and utilized in the subsequent Section \ref{subsec:benchmark_ppo}. 
This auxiliary reward adds a penalty term based on the ratio of the number of incorrect pixels to the total pixels, guiding the agent to learn in a direction that minimizes the number of pixels differing from the correct grid. 
Identifying a reward setting superior to this auxiliary reward setup, i.e., one that can be universally applied across all ARC tasks aware environment's action space (e.g., object-oriented operations), requires further research.

\subsection{Source Code}

Since the environments in ARCLE implemented based on Gymnasium~\citep{farama2023gymnasium} and are fully written in Python3, users who have used Gymnasium or its predecessor, OpenAI Gym~\citep{brockman2016openai}, can use it with familiarity.
ARCLE is released on GitHub\footnote{\url{\arclerepo}} under the terms of the Apache-2.0 License, as well as uploaded to the PyPI (Python Package Index), so the ARCLE can be easily installed by the \texttt{pip} command.\footnote{
$\texttt{\$ pip install arcle==0.2.5}$} Without modifying the source code, one can still create custom ARCLE-based environments by subclassing provided environments in ARCLE or wrapping with the wrapper classes. 
Please note that ARCLE is currently being continuously updated, so users may need to check the version. In this paper, our descriptions and experiments are based on version 0.2.5.

\subsection{API \& Sample Usage}
\usemintedstyle{vs}

\begin{figure}[!ht]
  \centering
  \begin{minipage}[t]{.48\linewidth}
    \captionof{lstlisting}{Basic Usage of an ARCLE environment (\texttt{O2ARCEnv}) with Gymnasium API. Action is randomly sampled.}
    \label{lst:arcle_sample_usage}
\begin{lstlisting}[language=Python]
import arcle
import gymnasium as gym

env = gym.make('ARCLE/O2ARCEnv', render_mode='ansi')
obs, info = env.reset(options={'adaptation': True})

for _ in range(1000):
    action = env.action_space.sample()
    obs, reward, term, trunc, info = env.step(action)
    if term or trunc:
        obs, info = env.reset()
\end{lstlisting}
  \end{minipage}\hfill
  \begin{minipage}[t]{.48\linewidth}
    \captionof{lstlisting}{BBox wrapping of the environment. Action is randomly sampled once, resulting in a 5-tuple. Continuing code from Listing \ref{lst:arcle_sample_usage}.}
    \label{lst:arcle_bbox_usage}
\begin{lstlisting}[language=Python]
from arcle.wrappers import BBoxWrapper

env = BBoxWrapper(env)

obs, info = env.reset()
action = env.action_space.sample()
print(action) # 5-tuple: (y1, x1, y2, x2, op)
\end{lstlisting}
  \end{minipage}
\end{figure}

Using Gymnasium API, an ARCLE environment can be created. 
Listing~\ref{lst:arcle_sample_usage} is the most basic usage of the ARCLE environment loop. 
In the code, the \texttt{O2ARCEnv} is created and its \texttt{reset} method is called to initialize the environment to the input grid state of a random ARC task. If \texttt{adaptation} is \texttt{True} in the reset options, the environment samples and initializes states and answers as a demonstration pair, otherwise, it initializes as a test input pair. 
Next, within a loop, the \texttt{sample} function from the Gymnasium API is executed to select a random action, and the \texttt{step} function applies this action to the current state. 
Finally, if the state reaches the correct solution, the \texttt{reset} function is executed again to start the loop over with a new ARC task.

An example of using a bounding box form for \texttt{selection} (in action space) instead of a raw binary mask, is shown in Listing \ref{lst:arcle_bbox_usage}. 
The environment is wrapped using a \texttt{BBoxWrapper} from ARCLE. 
As a result, the random action returned by the \texttt{sample} function consists of a tuple of five numbers, the first four values representing the bounding box of the \texttt{selection} and the last value specifies \texttt{operation}. 
While configuring \texttt{selection} as a raw binary mask for a grid of size $H \times W$ offers the advantage of allowing free-form object configurations, it also poses the problem of having a very large action space of $O(2^{HW})$. 
On the other hand, configuring \texttt{selection} as a bounding box simplifies it to four integers, reducing the action space to $O(H^2W^2)$, but it limits the shape of the object to a rectangle. 
This restriction is in place that necessitates the selection of background pixels when dealing with non-rectangular objects. However, ARCLE actions differentiate between zero-valued pixels, which are considered blank, and non-zero pixels. This distinction ensures that when the object is isolated from other pixels, there will be less overlap of irrelevant pixels by the background when object-oriented actions are applied.
\section{ARCLE Benchmarks}

This chapter explains the process through which an agent learns to solve synthetic tasks using the ARCLE environment. 
To ultimately solve ARC, the agent must acquire the ability to tackle unseen tasks through the learning process of tasks provided in the training dataset. 
We speculate that approaches like meta-RL, generative models (e.g. GFlowNet), and model-based RL algorithms (e.g. World Models) may be necessary to solve tasks not observed during training. 
As a preliminary step, we describe the initial results of learning an individual task. 
The PPO-based agent learns the input/output grid pairs presented in one of the ARC tasks. 
If a method can be designed for the agent to understand and learn from these tasks, we anticipate that it could be trained to solve unlearned tasks using the approaches mentioned above with this agent.

\subsection{Solving ARC with a given answer: handling the large discrete state-action space}
\label{subsec:benchmark_ppo}
While we expect ARCLE agents to be better at imitating the cognitive process of human problem-solving, training an RL agent for ARCLE itself additionally becomes a difficult challenge due to its large discrete state-action space. In this Section, we demonstrate the difficulty of obtaining highly performant agents within an ARCLE environment even when the state-action spaces are simplified and the answers are given, and we propose ARCLE-specific auxiliary loss functions and network architectures that can significantly improve agents' performance. Specifically, we use \texttt{operation}s of $0$--$9$ only with rectangular-shaped \texttt{selection} only (in a bounding box representation), and consequently, the sufficient information for decision making (i.e., the state $s$) becomes \texttt{(grid, grid\_dim, answer, answer\_dim)} as we additionally assume answers to be given. 
We expect the methods introduced here to be used to help train ARCLE agents for the original ARC, where the answers are not provided and state-action spaces are more complex.

\paragraph{Proximal Policy Optimization (PPO)} We employed the well-known PPO algorithm~\citep{schulman2017proximal} to train the agents to solve ARCLE with the answers given. Due to the poor generalization ability~\citep{kumar2021dr3} and learning instability of value-based RL algorithms, recently, PPO has been widely adopted for tuning large neural models~\citep{stiennon2020learning,ouyang2022training}. It is an on-policy policy-gradient algorithm that aims to perform a gradient update within the trust region. We gather trajectories and construct a dataset $\mathcal{D}=\{(s_i, a_i, R_i)\}_i$ consisting of state, action, and returns (sum of discounted rewards starting from the state). Then, the policy is updated according to the following losses ($\mathcal{L}=\mathcal{L}^{\text{Baseline}}+\mathcal{L}^{\text{PPO}}$) with samples from $\mathcal{D}$:
\begin{align*}
    \mathcal{L}^{\text{Baseline}}(\psi)&=\mathbb{E}_{\mathcal{D}}\left[
    \left(r - V_\psi(s)\right)^2
    \right]\\
    \mathcal{L}^{\text{PPO}}(\theta)&=\mathbb{E}_{\mathcal{D}}
    \left[
    \min\left(
    \frac{\pi_\theta(a|s)}{\pi_{\text{old}}(a|s)} \left(r-\texttt{sg}[V_{\psi}(s)]\right), \text{clip}\left(
    \frac{\pi_\theta(a|s)}{\pi_{\text{old}}(a|s)}, 1-\epsilon, 1+\epsilon
    \right)\left(R-\texttt{sg}[V_{\psi}(s)]\right)
    \right)
    \right],
\end{align*}
where $V_\psi$ is a value function that works as a baseline that reduces the gradient variance, $\pi_{\text{old}}$ is a policy used to gather the trajectories, and $\texttt{sg}[\cdot]$ is a stop-gradient operator. $r\in [-1, 0]$ is a reward from a dense reward function that penalizes the agent by the ratio of incorrect pixels of the next state.

\paragraph{State encoder} We used a shared state encoder for the policy $\pi_\theta$ and the baseline $V_\psi$ based on a Transformer encoder architecture~\citep{vaswani2017attention}. Each pixel of \texttt{grid} and \texttt{answer} is encoded as a token by taking a summation over corresponding position, color, and token type embeddings, where token type embedding informs whether it belongs to \texttt{grid} or \texttt{answer}. Depending on \texttt{grid\_dim} and \texttt{answer\_dim}, the tokens with an inactive pixel are masked so that it is not attended by other tokens. 
Each function gets its own special token(s) and feed-forward network to pass the extracted state feature from its token. The baseline $V_\psi$ use a single special token for its scalar output whereas the policy $\pi_\theta$ uses two or more tokens for representing both \texttt{operation} and \texttt{selection}, which will be detailed in Section \ref{sssec:cepa}.

In the following experiments, we only used tasks with \texttt{grid\_dim} and \texttt{answer\_dim} less than $5\times 5$ due to the computational demand of the current state encoder architecture. However, it can be alleviated by using more scalable architecture, e.g., a patch as a token instead of a pixel as a token~\citep{dosovitskiy2021image}. We experimented using two different settings, \textbf{(1)} a random setting where we use the randomly generated $5\times 5$ initial grid and goal, and \textbf{(2)} an ARC setting where we used initial grids and goals that are equal or smaller than $5\times 5$ from ARC tasks. In the random setting, we need to act precisely due to the vast number of different goals, whereas in the ARC setting, we need to adapt to various grid sizes.

\subsubsection{Learning better representation through auxiliary loss functions}
\label{sssec:alf}
\begin{figure}[t]
    \centering
    \includegraphics[width=\columnwidth]{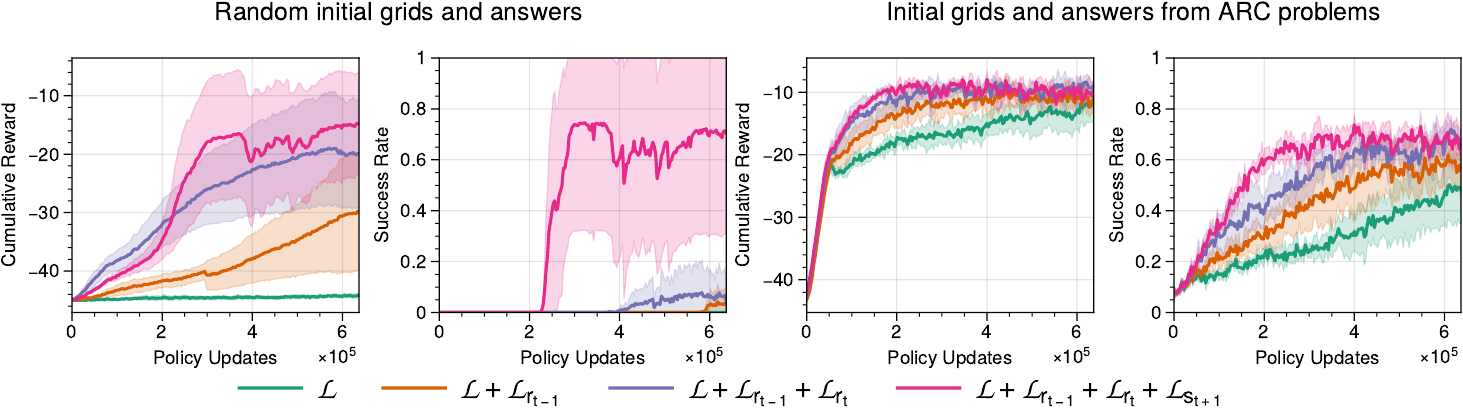}
    \caption{Performance of agents when various auxiliary losses are additionally used are shown. The experiment is repeated four times, and the shaded regions denote $95\%$ confidence intervals.}
    \label{fig:ppo_auxil}
\end{figure}

Using an auxiliary loss function to predict important information has been a widely used approach for better generalizable representation and faster training~\citep{jaderberg2016reinforcement,lample2017playing}. We experimented three different auxiliary losses, \textbf{(1)} $\mathcal{L}_{r_{t-1}}$ predicting the previous reward $r_{t-1}$ from the current state $s_t$, \textbf{(2)} $\mathcal{L}_{r_{t}}$ predicting the current reward $r_{t}$ from the current state-action $(s_t,a_t)$, and \textbf{(3)} $\mathcal{L}_{s_{t+1}}$ predicting the next state $s_{t+1}$ from the current state-action $(s_t,a_t)$. All three functions are deterministic, and they are highly informative as they are correlated to either the value function or the action-value function. For policy architecture, we used the color-equivariant policy architecture that will be detailed in Section \ref{sssec:cepa}.

While the first auxiliary loss $\mathcal{L}_{r_{t-1}}$ can be easily adopted by additionally training a feed-forward network on top of the extracted state feature from the special token of $V_\psi$, the other two auxiliary losses require the state-action feature that is not utilized in conventional PPO. We compute the state-action feature by performing a forward propagation again with additional action embedding tokens after sampling an action from a policy. The prediction for the loss $\mathcal{L}_{r_{t}}$ is done on top of the last action token that embeds \texttt{selection}, and the prediction for the loss $\mathcal{L}_{s_{t+1}}$ is done on top of tokens that represent each pixel of \texttt{grid}.

The results are reported in Figure \ref{fig:ppo_auxil}. Note that the vanilla PPO agent was not able to learn anything in the random setting despite the vastly simplified state-action space, demonstrating the difficulty of training an agent for ARCLE. 
While all of the experimented auxiliary losses improve the learning of the agent, it can be seen that adopting auxiliary features and adopting state-action feature-based auxiliary features make a significant difference in performance. Only with all three of these auxiliary losses, we were able to get three agents out of four that achieved a success rate larger than $95\%$ in the random setting. On the other hand, in the ARC setting, auxiliary losses were able to help, but their effect was relatively less dramatic.

\subsubsection{Non-factorizable policy architecture}
\label{sssec:cepa}
\begin{figure}[t]
    \centering
    \includegraphics[width=\columnwidth]{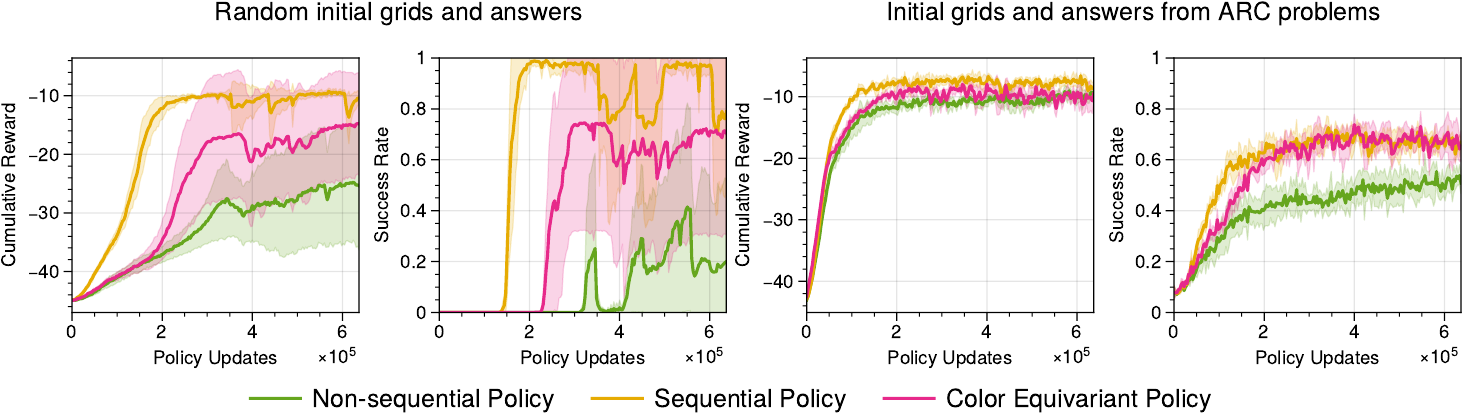}
    \caption{Performance of agents when equipped with different policy architectures. The experiment is repeated 4 times, and the shaded regions denote $95\%$ confidence intervals.}
    \label{fig:ppo_policy}
\end{figure}
 \def\ci{\perp\!\!\!\perp}
It can be observed that the two main components of the action space of ARCLE, \texttt{operation} and \texttt{selection}, are intertwined with each other and cannot be separately decided. For example, the optimal \texttt{selection} for coloring a pixel, or rotating an object will be completely different. This observation shows that the considerate choice of policy architecture is necessary, as conventional factorized policy assuming $(\texttt{operation} \ci \texttt{selection})~ | ~s$ will have limited expressivity in representing such complex multimodal distributions. For all experiments in this section, we used all three auxiliary losses introduced in Section \ref{sssec:alf}.

To demonstrate the expressivity of different policy architectures,  we experimented with the following three architecture types: \textbf{(1) Non-sequential policy} assumes $(\texttt{operation} \ci \texttt{selection})~ | ~s$. This policy is implemented by using two special tokens for \texttt{operation} and \texttt{selection} with two feed-forward networks on top of extracted features from those tokens. \textbf{(2) Sequential policy} does not assume conditional independence, by making the decision of \texttt{selection} dependent on sampled \texttt{operation}, similar to the RNN policy of \citet{vinyals2019grandmaster}. This policy requires two forward passes to sample an action, one for sampling \texttt{operation} from its special token and one for sampling \texttt{selection} from the token embedding the sampled \texttt{operation}, and therefore it is more computationally demanding. 

On the other hand, we also experimented \textbf{(3) Color-equivariant policy} that takes advantage of the ARCLE task that the same permutation of colors applied to the task and the policy coloring actions results in the equivalent task, i.e., the color equivariance. We can achieve color equivariance of the policy by using several special tokens for policy equal to the number of \texttt{operation} to represent them, and by setting a special token of color-related \texttt{operation} as a function of color embedding used to represent \texttt{grid}. We can then use two different feed-forward networks on top of extracted features of these \texttt{operation} tokens. One gives scalar output per token to be used as logits for deciding the \texttt{operation}. The other one is used to get the \texttt{operation}-specific \texttt{selection} on top of the sampled \texttt{operation} token. This policy only requires one forward pass with a few additional \texttt{operation} tokens, and it is computationally efficient compared to the sequential policy.

Figure \ref{fig:ppo_policy} summarizes the result. Overall, it can be observed that sequential policy and color equivariant policy outperform non-sequential policy, showing that conditional dependence is crucial for learning in ARCLE. Sequential policy shows more stable and faster learning compared to color equivariant policy in terms of policy updates. However, considering that sequential policy takes approximately $1.5$x training time and $2$x inference time, there is a clear trade-off and we can choose from two depending on the situation.

\begin{figure}[t]
    \centering
    \includegraphics[width=0.83\columnwidth]{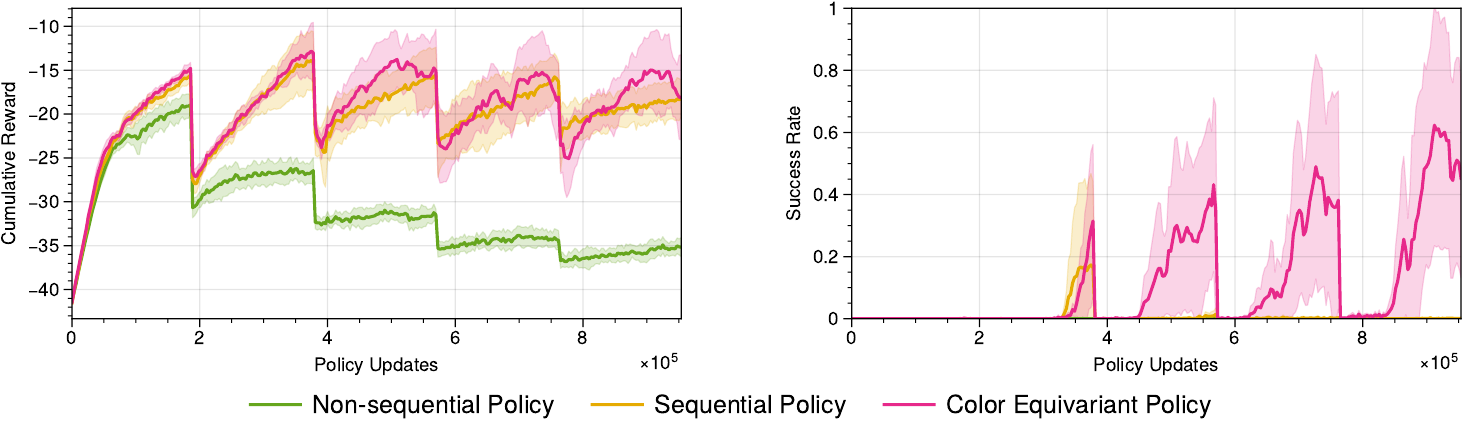}
    \caption{Performance of agents on continual RL task when equipped with different policy architectures. The experiment is repeated 4 times, and the shaded regions denote $95\%$ confidence intervals.}
    \label{fig:cont_policy}
\end{figure}

\subsubsection{ARCLE as a Continual RL Environment}
To address the inherent challenges of the ARCLE environment, it may be necessary to provide an agent with a curriculum. In such scenarios, an agent capable of continuously learning from a changing set of tasks would be beneficial. We conducted a continual RL experiment to demonstrate the robust learning capabilities of the proposed policy architectures in response to task changes. In this experiment, the initial grids and answers were randomly generated as before, but the number of colors used increased periodically—specifically, across five learning phases with 2, 4, 6, 8, and 10 colors respectively.

As depicted in Figure \ref{fig:cont_policy}, all agents experienced a significant drop in performance whenever the number of colors increased. Similar to what we have observed in Figure \ref{fig:ppo_policy}, the non-sequential policy cannot express the complicated dependencies between \texttt{operation} and \texttt{selection}, and is outperformed by the other two policy architectures. However, within the context of the continual RL experiment, the sequential policy was not able to adapt to the new sets of tasks and recorded a 0\% success rate after the second change in the environment. Conversely, the color equivariant policy demonstrated the ability to continuously improve its success rate, illustrating its rapid adaptability, which stems from its architectural design.

\subsection{Future Directions for RL Research in Solving ARC with ARCLE}

Based on Section~\ref{subsec:benchmark_ppo}, where we demonstrated the development and initial success of a PPO-based agent within ARCLE, this section aims to show future RL research in addressing the challenges. 
Inspired by \citet{chollet2019ARC}, we posit that an effective ARC solver must possess advanced abstraction and reasoning abilities. 
Thus, we propose several research directions using ARCLE (e.g. MAML, GFlowNet, and World Model), with more details in Appendix \ref{subsec:other_algorithms}.

\subsubsection{Meta-RL for Enhancing Reasoning Skills}
ARC is a multi-task few-shot learning problem: the whole dataset consists of multiple tasks, and each task has few demonstration pairs (supporting set) to infer the output of a test input (query set). 
ARCLE is in the identical problem but in the RL setting. 
To manage this, multi-task RL~\citep{wilson2007multi} or meta-RL~\citep{finn2017model} algorithms that foster an agent to experience over a task distribution could be applied. 
We have focused on developing ideas with meta-RL rather than multi-task RL as it gives a richer optimization. 
In this setting, the meta-training set and the meta-testing set are the training and evaluation sets given in the ARC dataset. 

Meta-RL algorithms on ARCLE should be capable of outputting an RL algorithm that rapidly reasons and produces a policy for each ARC task, without exhaustive searching over actions. 
The policy is trained to generate valid trajectories from the input to the output grids simultaneously on multiple demonstration pairs in an ARC task. 
Then the policy is applied to the test input grid to generate output. 
Therefore, Meta-RL endows agents with essential reasoning skills for ARC's diverse tasks, enabling them to quickly adapt to new tasks by autonomously developing learning strategies. 
Integrating Meta-RL with ARCLE opens new pathways for researchers to devise techniques that allow AI to effectively generalize learning across various tasks, thus embodying the `learning to learn' principle.

\subsubsection{Generative Models as Surrogates for Reasoning}
Generative models, particularly GFlowNet~\citep{bengio2023gflownet}, offer a novel approach to tackling the reasoning challenges presented by ARC.
While an agent is equally given a set of grid operations on the ARCLE, many possible trajectories can lead to a correct answer for an ARC task. 
Moreover, among demonstration pairs in an ARC task, the detailed trajectories for each pair are varied, as each pair has its own input grid. 
GFlowNet establishes its policy as a generative model that enables the sampling of actions from it, and the probability of sampling is proportional to the reward-driven objective.
Therefore, GFlowNet benefits from not only learning a posterior distribution to include high-reward modes but also from searching multiple modes of a solution space by leveraging probabilistic reasoning to generate diverse possible solutions, in the form of a directed acyclic graph (DAG). 
This supports a GFlowNet policy to solve the demonstration pairs in one ARC task, although its input grids are different from (but possess the same rule) one another.
Moreover, its ability to identify multiple viable solutions for individual ARC tasks underscores its utility for data augmentation with correct solutions, further enhancing its value as a research tool in this domain.

\subsubsection{Model-Based RL for Abstraction Skills}
Encoding the demonstration pairs based on the core knowledge is a crucial point in establishing a plan to solve an ARC task. 
Model-based RL, particularly World Models might be a solution to support abstraction in tackling the ARC pairs. 
Among the ARC tasks, there are dissimilar common rules over all pairs in a task, although, there are a few categories of core knowledge that a common rule in each task be derived from. 
Objectness, goal-directness, arithmetic, geometric, and topology are part of them~\citep{chollet2019ARC}, and these can be infused in ARCLE's actions, like \texttt{Move} and \texttt{Rotate} operations. 
Since World Models internalize the environment transitions caused by ARCLE's actions to learn an agent on its simulation, it would learn a joint representation of ARCLE's grid pair and actions (containing core knowledge). 
It encourages a controller in World Model agent to utilize flexible neural representation, rather than hard-coded operations in ARCLE to simulate. 
In short, World Models would provide neural abstraction skills of the pairs and operations in ARCLE, which supports a controller to search a rule efficiently on the flexible representation.
Hence, developing agents that can construct and utilize these models is a step towards equipping them with the necessary abstraction skills for handling both trained and untrained tasks.

\subsubsection{Further Research Questions}
Several research questions would advance while tackling ARC with ARCLE. 
First, the ARC task does not possess an explicit task distribution since individual ARC tasks include a unique rule and the current ARC dataset has only a finite 800 training and evaluation tasks. 
Categorizing ARC tasks correspondingly to the core knowledge~\citep{moskvichev2023conceptARC} and parameterizing tasks in each category similarly to XLand~\citep{bauer2023human} would be a worthwhile topic that reinforces meta-RL and multi-task learning more promising methods to solve ARC with broader tasks.

Next, ARCLE's action space consists of two sub-action spaces: an integer \texttt{operation} and a discrete binary mask \texttt{selection}. 
Handling with \texttt{selection} might entail an exponential size of search space: in particular, the size of DAG with the GFlowNet approaches grows enormously to degrade the efficiency of figuring out the correct output grid. 
One probable setting is that we utilized a sequential policy in \ref{sssec:cepa}, maintaining two networks that produce \texttt{operation} and \texttt{selection} hierarchically. 
Then this GFlowNet may maintain a DAG of sampling \texttt{operation} only by considering ARCLE as a probabilistic environment. 
However, the validity of this method is indeterministic.

Lastly, one might doubt the necessity of World Models in solving ARC to provide richer abstraction. 
The reason would be that training agents directly in the environment is more straightforward since the environment’s dynamics are deterministic, rather than learning the World Models. 
Nevertheless, in ARC and ARCLE, there is a significant amount of auxiliary information to abstract more than a transition of observation: object information and their topology, symmetry, and so forth. 
Previous studies have shown that it can capture information such as a hidden gravity parameter in an environment~\citep{reale2022disentangled}, and it is expected that additional useful information for ARC can be extracted as well. 
One open question brought here is what information a World Model can extract for ARC. In particular, whether it can disentangle and extract information common to every ARC task (e.g., state transition) and task-specific priors (e.g., objectness) in an interpretable form is a research question for the future.
\section{Conclusion}
In this paper, we introduced ARCLE, an RL environment designed for the ARC benchmark, using the Gymnasium library for direct engagement with ARC's challenges. 
Our development and application of a PPO-based agent, enhanced with auxiliary losses and non-factorizable policies, have demonstrated ARCLE's possibility of learning and performance improvements in addressing ARC tasks. 
In detail, auxiliary losses improved learning outcomes, especially evident in random settings where the comprehensive application of all proposed strategies yielded the best performance. 
The success rate in these settings, and the superior outcomes from applying sequential and color-equivalent policies, underline the importance of strategic \texttt{operation} and \texttt{selection} processes.

These experimental results show advanced RL methodologies—such as meta-RL, generative models, and model-based RL—to further enhance AI's reasoning and abstraction abilities. 
Specifically, meta-RL offers the potential to refine AI's reasoning skills by enabling adaptive learning strategies across varied tasks, suggesting a path toward more generalized intelligence. 
Generative models, by simulating complex reasoning processes, could serve as links between data and sophisticated decision-making in ARC. 
Model-based RL model could strengthen AI's ability to distill and apply abstract concepts from complex inputs. 
Thus, further research using ARCLE could elevate AI's learning strategies and expand the boundaries of its current capabilities.
We invite the RL community to engage with ARCLE not just to solve ARC but to contribute to the broader endeavor of advancing AI research. 
Through such collaborative efforts, we can unlock new horizons in AI's ability to learn, reason, and abstract, marking significant progress in the field.

\subsubsection*{Acknowledgments}
This work was supported by the IITP (RS-2023-00216011, No. 2022-0-00311), the National Research Foundation (RS-2023-00240062), Artificial Intelligence Graduate School Program (No. 2019-0-00079, No. 2019-0-01842), and GIST (AI-based Research Scientist Project) funded by the Ministry of Science and ICT, Korea. 

\bibliography{main}
\bibliographystyle{collas2024_conference}

\newpage
\appendix
\section{Appendix}

\subsection{Abstraction and Reasoning Corpus (ARC)}
\label{subsec:arc}

With increasing interest in human-like AI, attempts to measure intelligence are being made. However, how can intelligence be quantified? Previous research has defined intelligence as the ability to solve various types of problems with limited data and experience~\citep{chollet2019ARC}.

\begin{align*}
    \textit{I}^\textit{opt}_\textit{IS, scope}=\underset{T \in \textit{scope}}{\textit{Avg}}[\omega_{T, \Theta}\cdot\Theta\underset{\textit{C} \in \textit{Cur}_T^{opt}}{\sum}[\textit{P}_\textit{C}\cdot\frac{GD_{IS,T,C}^\Theta}{P_{IS,T}+E_{IS,T,C}^{\Theta}}]]
\end{align*}

In the above equation, $\omega_{T,\Theta} \cdot \Theta$ represents the weights, $C$ denotes a single curriculum (training data), $P_C$ is the probability of the curriculum occurring, $GD_{IS,T,C}^\Theta$ signifies the generalization difficulty of solved problems, $P_{IS,T}$ represents prior knowledge, and $E_{IS,T,C}^{\Theta}$ denotes the model's experience. Note that intelligence gets bigger when prior knowledge and experience get smaller and generalization difficulty gets bigger, just as the definition of intelligence proposed above. Therefore, it was argued that benchmarks for evaluating intelligence must meet three criteria: (1) solvable with limited prior knowledge alone, (2) composed of diverse problem types, and (3) quantitatively measurable.

ARC is a benchmark proposed to measure the intellectual capabilities of computers quantitatively. Each task in ARC consists of $2$--$5$ demo pairs with both inputs and outputs given, along with one test input grid. The goal is to infer the solution to the test input grid by deducing rules from the examples. Demo inputs and outputs can vary in size from a minimum of $1\times1$ grid to a maximum of $30\times30$ grid, with each grid capable of being colored with $10$ different colors. One other property of ARC is that it is solvable with just four types of prior knowledge: objectness, goal-directedness, counting, and geometry and topology~\citep{chollet2019ARC}. Thus, ARC is considered a fair intelligence assessment scale because it requires \textbf{relatively simple rules and limited prior knowledge} to solve tasks, while also necessitating the inference of \textbf{various rules} and enabling \textbf{numerical evaluation} of whether a task can be solved or not.

One of the key features of ARC is its requirement for high levels of abstraction and reasoning compared to other benchmarks. Previous research comparing benchmarks for evaluating visual reasoning abilities noted that ARC stands out because it requires understanding abstract images and rules, evaluates by generating answers, and can present unseen tasks~\citep{malkinski2023review}. Due to these characteristics, as of $2023$, state-of-the-art models demonstrate approximately 30\% accuracy~\citep{johnson2021fast}. When compared to the fact that human performance is around $80$\%, it becomes evident how challenging the benchmark is. This is why ARC gathers attention in the pursuit of human-like AI research.

\begin{table}[!htbp]
    \caption{Alignment of Abstract Visual Reasoning tasks with its taxonomy~\citep{malkinski2023review}. The problems and their corresponding benchmarks are cataloged under the following four dimensions of the taxonomy: Input shapes, Hidden rules, Target tasks, and Specific challenges.}
    \centering
    \renewcommand{\arraystretch}{1.5}
    \resizebox{\textwidth}{!}
    {\begin{tabular}{l|cc|cc|cc|cc}
    \toprule
        \textbf{Dataset} & \makecell{\textbf{Geometric} \\ \textbf{Shapes}} & \makecell{\textbf{Abstract}\\ \textbf{Shapes}} & \makecell{\textbf{Explicit} \\ \textbf{Rules}} & \makecell{\textbf{Abstract} \\ \textbf{Rules}} & \textbf{Classify} & \textbf{Generate} & \makecell{\textbf{Domain} \\ \textbf{Transfer}} & \textbf{Extrapolate}\\ \midrule
        \rowcolor{yellow}
        \textbf{ARC}~\citep{chollet2019ARC} & ~ & \checkmark & ~ & \checkmark & ~ & \checkmark & \checkmark & \checkmark \\ \midrule
        \textbf{Sandia} ~\citep{matzen2010recreating} & \checkmark & ~ & \checkmark & ~ & \checkmark & ~ & ~ & ~ \\ 
        \textbf{Synthetic} ~\citep{wang2015automatic}& \checkmark & ~ & \checkmark & ~ & \checkmark & ~ & ~ & ~ \\ 
        \textbf{G-set}~\citep{mandziuk2019deepiq} & \checkmark & ~ & \checkmark & ~ & \checkmark & ~ & ~ & ~ \\ 
        \textbf{RAVEN}~\citep{zhang2019raven} & \checkmark & ~ & \checkmark & ~ & \checkmark & ~ & \checkmark & ~ \\ 
        \textbf{PGM}~\citep{barrett2018measuring} & \checkmark & ~ & \checkmark & ~ & \checkmark & ~ & \checkmark & \checkmark \\ 
        \textbf{Hill et al.}~\citep{hill2019learning} & \checkmark & ~ & \checkmark & ~ & \checkmark & ~ & \checkmark & \checkmark \\
        \textbf{G1-set}~\citep{mandziuk2019deepiq} & \checkmark & ~ & \checkmark & ~ & \checkmark & ~ & \checkmark & ~ \\  
        \textbf{S1-set}~\citep{mandziuk2019deepiq} & \checkmark & ~ & \checkmark & ~ & \checkmark & ~ & \checkmark & ~ \\
        \textbf{MNS}~\citep{zhang2020machine} & \checkmark & ~ & \checkmark & ~ & \checkmark & ~ & ~ & ~ \\ 
        \textbf{VAEC}~\citep{webb2020learning} & \checkmark & ~ & \checkmark & ~ & \checkmark & ~ & ~ & \checkmark \\ 
        \textbf{DOPT}~\citep{webb2020learning} & \checkmark & ~ & \checkmark & ~ & \checkmark & ~ & ~ & \checkmark \\ \bottomrule
    \end{tabular}}
\end{table}

\newpage
\subsection{Significance of Solving ARC using Reinforcement Learning}
\label{subsec:RL_significance}

Just as ARC has significant implications for the field of reinforcement learning (RL), RL also holds great importance for the ARC and Artificial General Intelligence (AGI) research areas. This is because RL methodology 1) has characteristics suitable for solving ARC compared to other approaches, 2) has a high possibility of leading to research on human reasoning, which is the aim of ARC, and 3) facilitates the utilization of research resources that have been employed in other fields.

\paragraph{The suitability of RL for solving the ARC}
RL possesses suitable characteristics for solving ARC. ARC can be understood as a program synthesis benchmark that involves composing complex solutions from simple skills~\citep{chollet2019ARC}. Previous attempts to solve ARC support this claim. Conventional deep learning approaches like autoencoders show performance below 10\%~\citep{veldkamp2023solving}, as they are specialized in learning a single skill for solving a specific task. Similarly, large language models that have achieved success across various domains also exhibit around 15\% performance~\citep{mirchandani2023large}, likely due to their limitation in finding combinations of step-by-step skills. The highest performance has been achieved by program synthesis methods that search for combinations of human-designed Domain Specific Languages (DSLs)~\citep{Hodel2023arcathon}. While these results support that ARC indeed requires program synthesis components, the current methods using manually designed DSLs are limited in their vulnerability to unseen tasks and human bias. On the other hand, RL is a research field specialized in solving complex problems by composing simple actions. Methods like MuZero~\citep{schrittwieser2020mastering} have successfully found effective action combinations in environments like the game of Go. The strong compositional capability makes RL more suitable for solving ARC than other approaches.

\paragraph{Potential for expanding into human reasoning research}

RL approach to solving ARC is expected to provide a significant foundation for theoretical inquiries into AGI, as RL methodologies are similar to how humans solve tasks. While there have been several attempts to solve ARC, existing methods differ from the strategies typically employed by intelligent agents for general problem-solving. Recent research methods, spearheaded by large language models, have been argued to diverge from human reasoning processes~\citep{mitchell2024comparing}. In contrast, RL has consistently yielded findings suggesting its biological similarity to human reasoning processes~\citep{matsuo2022deep, stachenfeld2017hippocampus}. Additionally, the use of intuitive rewards and policies allows for a transparent examination and approach to tasks compared to other machine learning methods. This transparency in understanding how actual reasoning occurs is one of the crucial characteristics that enables research into the nature of reasoning. Therefore, research on ARC through RL will not only aim to solve the tasks but also present an opportunity to gain insights into how humans reason from a biological perspective.

\paragraph{Acquisition of diverse research resources}

ARC is a benchmark created in 2019, and its solutions are currently under research. Supporting an RL environment that can solve ARC has the effect of easily introducing resources from other machine learning research fields into ARC. RL allows the application of deep learning models and techniques developed in other fields such as vision and natural language processing as its policy function, making it advantageous for utilizing existing resources. These characteristics will further facilitate the application of recently spotlighted methods such as meta-learning, continual learning, and multi-task learning to ARC. Consequently, rather than being limited to finding RL-specific solutions, it provides a great opportunity to test various methodologies and resources together.

Despite these advantages, efforts to tackle ARC tasks using RL have been limited, mainly due to a lack of appropriate RL environments. ARCLE maintains the inherent difficulty of ARC that requires solving general tasks with minimal prior knowledge and experience while preserving the strength of the RL approach in its similarity to human reasoning by incorporating the actions humans use when solving ARC. Furthermore, the action space consisting of low-level actions and the environment that allows for various variations offer high potential for utilization in meta-learning and continual learning research. Owing to these characteristics, ARCLE will not only contribute to the field of RL research but also make significant contributions to research on ARC and reasoning intelligence.

\newpage
\subsection{Object-Oriented ARC (O2ARC) Web Interface}
\label{subsec:o2arc}

Object-Oriented ARC (O2ARC) is a web interface that allows humans to directly solve ARC tasks and collect the process of solving them~\citep{kim2022playgrounds, shim2024o2arc}\footnote{\url{https://o2arc.com}}, an improvement upon the initial testing web interface developed by \citet{chollet2019testinginterface}. Initially, Chollet's testing web interface featured only a basic version involving coloring. However, the version of O2ARC has progressively improved to include object-oriented actions such as movement, rotation, and mirroring. Since actions represent the most intuitive low-level actions conceivable by humans, the sequence of actions (traces) could be used as a dataset reflecting human cognitive processes. The most recent version of O2ARC allows for the collection of traces solved by humans and also includes the ability to create tasks directly, thus evolving into a tool that can aid in the development of general artificial intelligence capable of mimicking human cognitive processes using ARC. The dataset collecting human traces is valuable in the research and development of artificial intelligence capable of human-like thinking.

Previous research has been conducted on whether learning from human traces can solve tasks~\citep{park2023unraveling}. This research demonstrated that with a sufficient number of traces solved by humans, it becomes feasible to solve ARC tasks by reflecting human solutions, thereby showcasing the potential of offline reinforcement learning. In line with this, ARCLE has been developed by integrating the actions of O2ARC to investigate whether an agent can address ARC tasks like human thinking. While O2ARC has one of its strengths in collecting human traces, ARCLE has the advantage of being able to train agents using the actions same as O2ARC. Therefore, ARCLE can be seen as having transformed O2ARC into a reinforcement learning environment for solving ARC tasks by agents. In Section \ref{subsec:other_algorithms}, we explore the possibility of using alternative RL algorithms to solve ARC tasks.

\begin{figure}[htbp!]
    \centering
    \includegraphics[width=0.95\columnwidth]{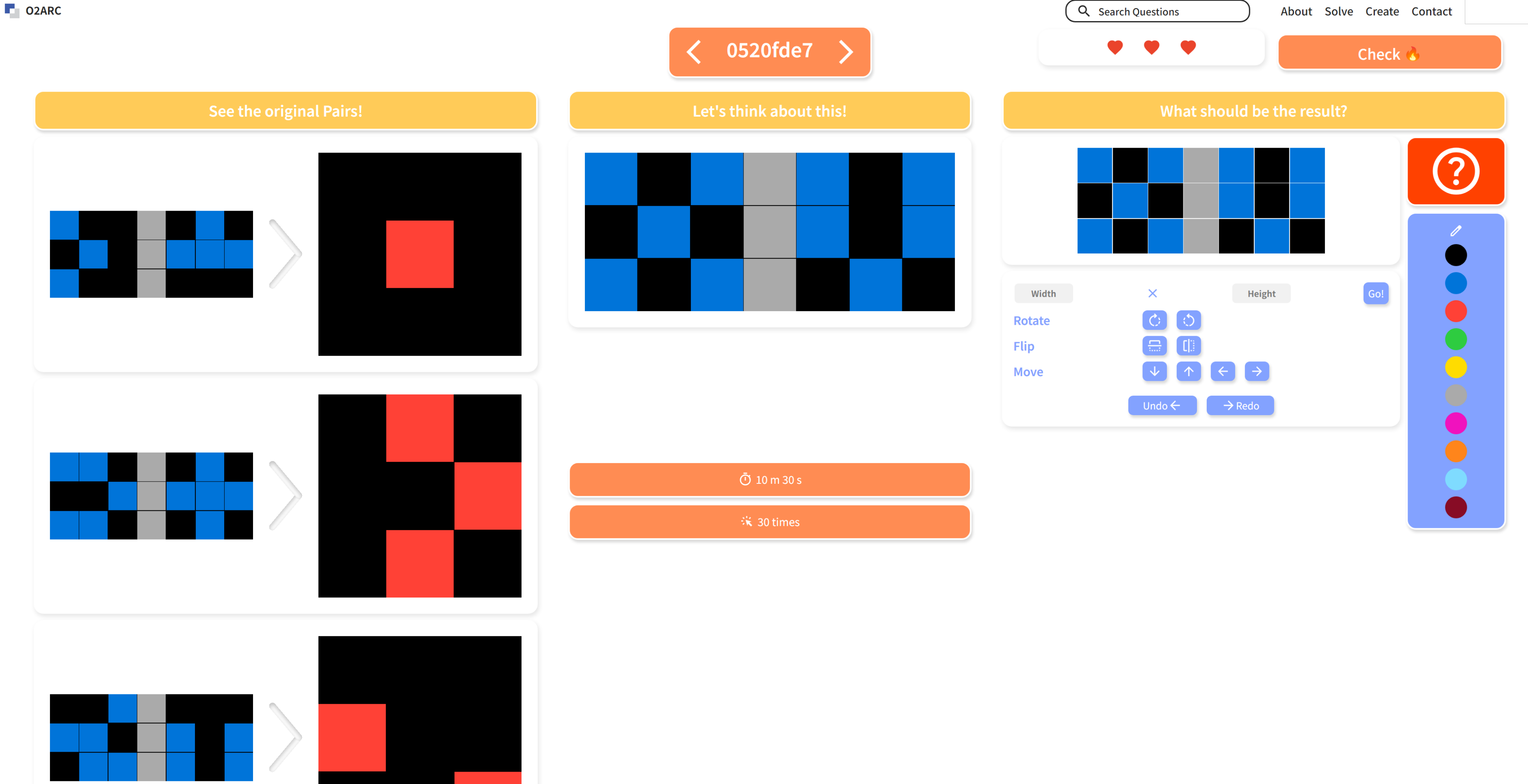}
    \caption{O2ARC Solve page. The left side shows demo input and output grid pairs, the center demonstrates the test input grid, and the right side consists of the result grid and available actions.}
    \label{fig:o2arc_solve}
\end{figure}

Figure \ref{fig:o2arc_solve} presents the interface for solving ARC tasks in O2ARC. As depicted, The left side part ``See the original pairs" displays demo pairs corresponding to the task, while the center provides the test input grid for the task. Users infer common rules from these examples to guess the appropriate answer for the given input. The right side features a grid space labeled ``What should be the result" where users input their answers. Additionally, a palette on the far right offers a selection of 10 colors for ARC. Users can use O2ARC's functionalities to color pixels, select objects, and perform actions such as rotation or copying and pasting. If necessary, they can input integer numbers into width and height cells to resize and submit the correct answer. The detail about the actions of O2ARC and ARCLE is explained in Section \ref{subsec:action_description}.

\newpage
\subsection{Every operation including ARCLE}
\label{subsec:action_description}

\begin{figure}[htbp!]
    \centering
    \includegraphics[width=\columnwidth]{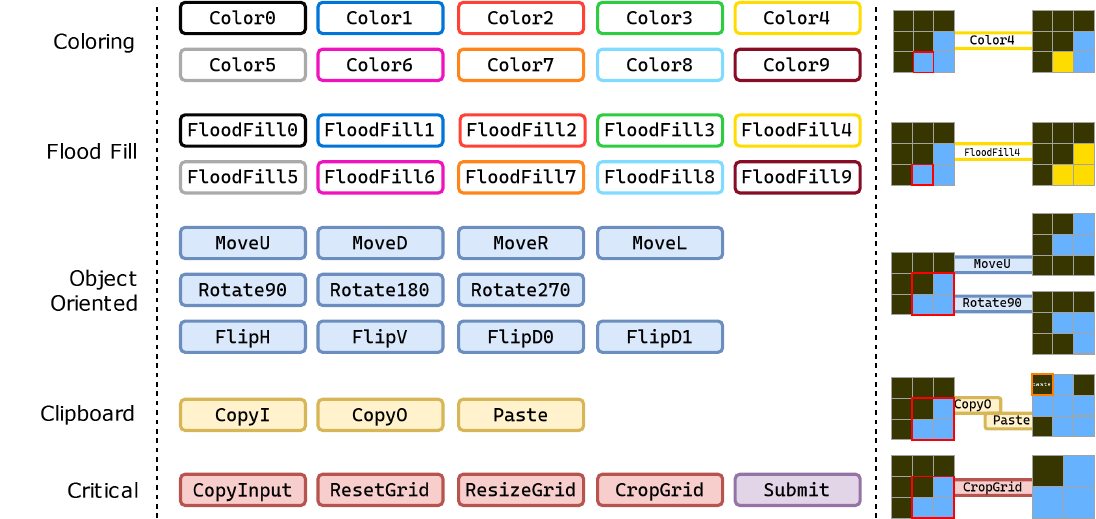}
    \caption{Every operation in ARCLE (version 0.2.5). }
    \label{fig:arcle_all_ops}
\end{figure}

Figure \ref{fig:arcle_all_ops} shows all the \texttt{operations} currently present in ARCLE.
As we mentioned before, the environments in ARCLE support only a subset of \texttt{operation}, tailored to their respective purposes. 
For instance, the simplest environment, \texttt{ARCRawEnv}, includes every Coloring \texttt{operation} and two Criticals (\texttt{ResizeGrid} and \texttt{Submit}).

The operations in the ARCLE \textbf{actions} component are categorized into five groups: Coloring, Flood Fill, Object-Oriented, Clipboard, and Critical operations. 
\textbf{Coloring} operations (\texttt{Color0}--\texttt{Color9}) change the color of selected pixels specified in \texttt{selection}. \textbf{Flood Fill} (\texttt{FloodFill0}--\texttt{FloodFill9}) alters the color of multiple pixels sharing the same color of \texttt{selection} pixels, as depicted in the right side of Figure \ref{fig:arcle_all_ops}. Two groups each consist of ten operations corresponding to ten distinct colors, defined in the ARC.

\textbf{Object-oriented} operations refers to \texttt{Move}, \texttt{Rotate}, and \texttt{Flip} added to O2ARC, by regarding pixels within the grid as object(s). \texttt{MoveU}, \texttt{MoveD}, \texttt{MoveR}, and \texttt{MoveL} shifts selected pixels in the specified direction (up, down, right, left). \texttt{Rotate90}, \texttt{Rotate180}, \texttt{Rotate270} turn them by 90, 180, or 270 degrees counterclockwise respectively. \texttt{FlipH}, \texttt{FlipV}, \texttt{FlipD0}, \texttt{FlipD1}  mirror object(s) horizontally, vertically, or diagonally (major and minor diagonal). If the \texttt{selection} is non-rectangular, object-oriented operations first calculate the bounding box of \texttt{selection} and operate in advance. Black pixels (zero-valued) are considered blank or transparent pixels, therefore those pixels do not affect other pixels by overlapping while performing sequences of object-oriented operations.

\textbf{Clipboard} operations involves the clipboard of the state, \texttt{clip}, copying from selected and pasting to the grid. \texttt{CopyI} copies pixels from test input grid (\texttt{input} in state space) specified by \texttt{selection} of action. On the other hand, \texttt{CopyO} copies pixels from the editing grid (\texttt{grid} of the state). \texttt{Paste} overlay the pixels of the \texttt{clip} state, on the editing grid (\texttt{grid} of the state). Pasting location is specified by the left-top corner of the bounding box of \texttt{selection} binary array.

\textbf{Critical} operations significantly modify the grid; every operation affects the editing grid by replacing the input grid (\texttt{CopyInput}), clearing its pixels to black (\texttt{ResetGrid}), changing the grid's size (\texttt{ResizeGrid}), and cropping the grid (\texttt{CropGrid}). \texttt{ResizeGrid} changes the grid's height and width as indices of bottom-right pixels of the \texttt{selection}. \texttt{CropGrid} directly puts selected pixels into \texttt{grid}, with resizing \texttt{grid} as a bounding box of selected pixels. Lastly, \texttt{Submit} operation, highlighted in purple, submits the current editing grid to compare with the answer of the assigned ARC task.

\newpage
\subsection{Two-Layer Mechanism of Object-Oriented Actions}
\label{subsec:technical_details}

\begin{figure}[htbp!]
    \centering
    \includegraphics[width=\columnwidth]{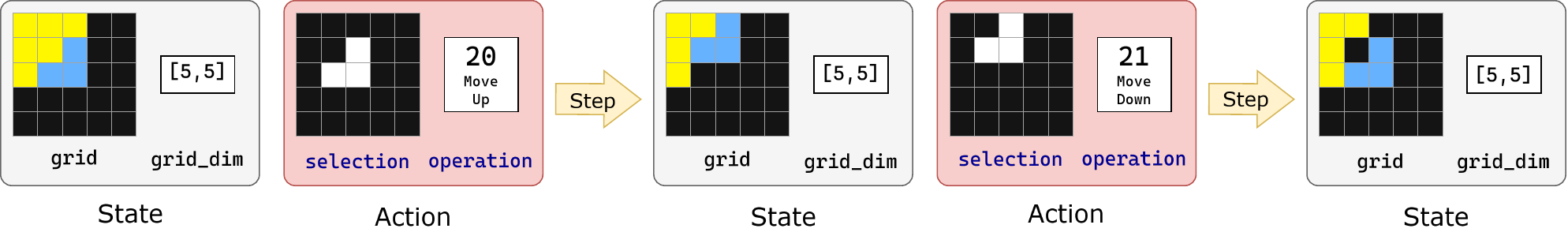}
    \caption{The edge case with consecutive \texttt{Move} actions.}
    \label{fig:arcle_one_layer_mechanism}
\end{figure}

The actions implemented in ARCLE, such as \texttt{Move}, \texttt{Rotate}, \texttt{Flip}, are object-oriented actions that act on the pixels selected by the agent, i.e., the objects. 
Simplifying the implementation of these actions to merely move the selected pixels could lead to issues as illustrated in Figure \ref{fig:arcle_one_layer_mechanism}.

Figure \ref{fig:arcle_one_layer_mechanism} demonstrates the issue with a simplistic implementation of the \texttt{Move} action, depicted through the process of an agent performing two \texttt{Move} actions. 
In the first \texttt{Move} action, the gray pixels (object) included in the \texttt{selection} move upwards, overlapping with the yellow pixels directly above the object, and the pixels vacated by the object's movement are filled with the background color, black. 
When the gray object is moved back down in the second \texttt{Move} action, the pixels previously painted with the background color during the first move overlap with the object, and similarly, the vacated pixels are filled with the background color, black. 
As a result, the yellow pixels that overlapped with the object during the first \texttt{Move} action are changed to the background color, black. 
Thus, a simple implementation of object-oriented actions can lead to the disappearance of pixels that overlap as the object moves.

\begin{figure}[htbp!]
    \centering
    \includegraphics[width=\columnwidth]{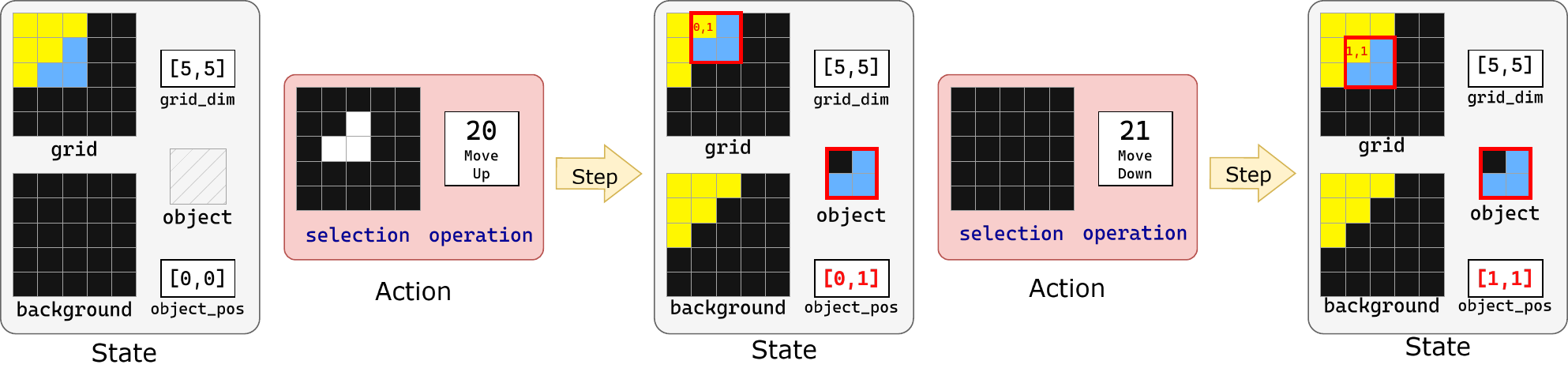}
    \caption{The edge case with serial two-layer mechanism \texttt{Move} actions.}
    \label{fig:arcle_two_layer_mechanism}
\end{figure}

To prevent information loss during the movement of objects, we implemented ARCLE's object-oriented actions using a two-layer mechanism. 
This approach is inspired by the way people typically lift and move objects, dividing the state space's \texttt{grid} into an \texttt{object} layer, which includes currently selected pixels, and a \texttt{background} layer, which comprises the rest of the pixels. 
Actions are performed on the \texttt{object} layer, which is then placed over the \texttt{background} layer to create the final \texttt{grid}. 
This two-layer mechanism for object-oriented actions utilizes variables stored in the \texttt{object\_states} dictionary within the state space, and the detailed operation process is as Figure \ref{fig:arcle_two_layer_mechanism}.

At the start of an object-oriented action, the \texttt{active} variable in the dictionary is set to $1$, and the pixels designated by the agent's \texttt{selection} in the action space are stored in \texttt{object}, while the rest are stored in \texttt{background}. 
The top-left coordinate of the bounding box surrounding the \texttt{object} is saved in \texttt{object\_pos}. 
If the action performed is not object-oriented, the \texttt{active} variable is set to $0$, and the \texttt{object} is reset. 
When \texttt{active} is $1$, meaning an object-oriented action was performed previously, and the agent performs another object-oriented action, only \texttt{object}, \texttt{object\_pos}, or \texttt{object\_dim} change depending on the type of action, while \texttt{background} remains unchanged. 
Upon completing an object-oriented action, \texttt{background} and \texttt{object} merge using the information from \texttt{object\_pos}, and the result is stored in \texttt{grid}. 
For example, in the two-layer mechanism, the \texttt{Move} action is implemented such that only the location of the \texttt{object} changes, altering only the value of \texttt{object\_pos} during the action, while the rest of the variables in the dictionary, like \texttt{object} or \texttt{object\_dim}, do not change.

\newpage
\subsection{Other RL Algorithms Trials}
\label{subsec:other_algorithms}

For future convenience of application, we provide three types of reinforcement learning models using ARCLE. These include Meta-RL, which has shown success in solving various types of tasks, GFlowNet, which excels in exploring wide search spaces, and World Model, which has strengths in analyzing complex domains. Such applications demonstrate that ARCLE can be utilized not only for standard reinforcement learning algorithms that demand strict reward and action specifications.

\subsubsection{Meta-RL Algorithms}
\label{subsubsec:meta-rl}

\paragraph{Model-Agnostic Meta-Learning}

\begin{figure}[htbp!]
    \centering
    \includegraphics[width=\textwidth]{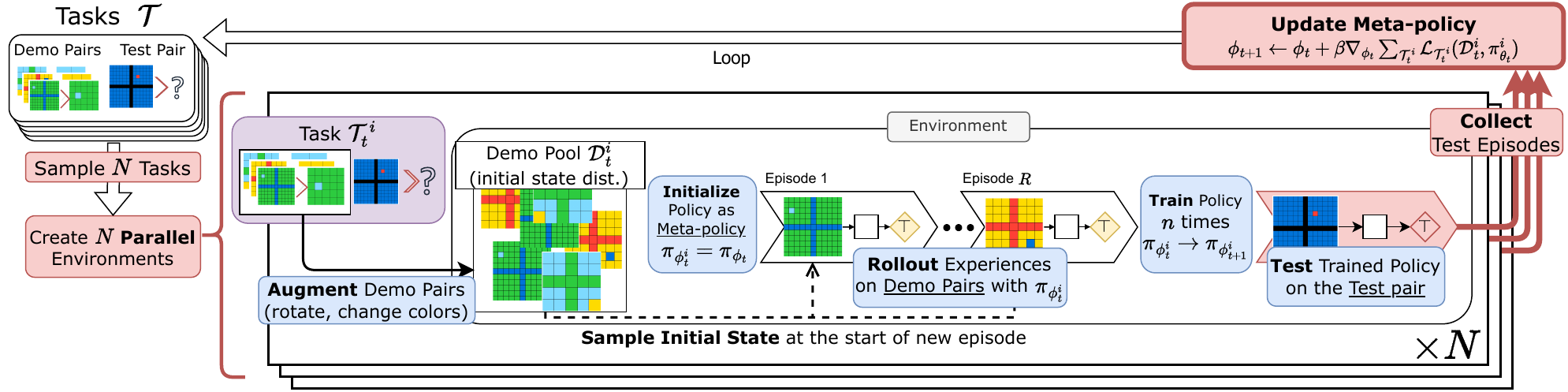}
    \caption{Learning Procedure of MAML integrated with ARCLE to solve ARC.}
    \label{fig:maml_architecture}
\end{figure}

We introduce the architecture of MAML~\citep{finn2017model}, the most famous parameterized policy gradient (PPG)~\citep{beck2023survey} meta-RL algorithm, integrated with ARCLE follows the training process as described in Figure \ref{fig:maml_architecture}. 
At time $t$, when a subset of the various tasks ($\mathcal{T}_1, \cdots, \mathcal{T}_N$) stored in ARCLE is sampled, it generates augmented demo pairs $\mathcal{D}^{i}_{t}$ from the demo pair available for task $\mathcal{T}^{i}_{t}$. 
Each inner loop holds a policy $ \pi_{\phi_{t}}$ created with the current initial parameter $\phi_{t}$, and trains by sampling and rolling out augmented demos ($\mathcal{D}^{i}_{t}$). 
The updated parameter $\phi^{i}_{t+1}$ during the training process is obtained as per the following Equation \ref{eq:maml_inner_loop}.

\begin{equation}
\label{eq:maml_inner_loop}
\phi^{i}_{t+1} = \phi_{t} - \alpha \nabla_{\phi_{t}}\mathcal{L}_{\mathcal{T}_t^i}(\pi_{\phi_t})
\end{equation}

Afterward, in the stage known as meta-testing, attempts are made to solve each task using the policy ($ \pi_{\phi_{t}}$) from each inner loop. 
The outcomes of these attempts are then utilized to update the parameters of the meta policy, by the Equation \ref{eq:maml_outer_loop}.

\begin{equation}
\label{eq:maml_outer_loop}
\phi_{t+1} = \phi_{t} - \beta \nabla_{\phi_{t}}\sum\limits_{\mathcal{T}^i_t \sim p(\mathcal{T})} \mathcal{L}_{\mathcal{T}_t^i}(\pi_{\phi_{t+1}^i})
\end{equation}

\paragraph{Expected usage}
Meta-RL is a type of meta-learning where conventional RL is used in the inner loop. 
This approach allows us to view the ARC task from a meta-learning perspective, where the goal is to train models on tasks with minimal information and then evaluate them on new tasks. 
If we use models capable of effectively learning an individual ARC task in the inner loop, like the PPO-based model proposed in this study, meta-learning will enable these models to learn how to solve a variety of ARC tasks. 
Consequently, such trained models will also gain the ability to solve untrained tasks.

However, there are concerns that the inner loop of meta-RL may struggle to learn effectively due to ARC tasks being composed of very limited information (3--5 demo grid pairs and a test input grid). 
Additionally, the vast action space of ARC could also hinder learning. 
To address these issues, RL techniques such as offline meta-RL or hindsight experience replay might be necessary. 
Particularly for offline meta-RL, a buffer containing a large number of trial records might be required. Collecting such records through an interface like O2ARC or utilizing methods such as data augmentation could be potential solutions.

\newpage
\subsubsection{Generative Flow Network}
\label{subsubsec:gflownet}

\paragraph{Generative Flow Network}
Generative Flow Network (GFlowNet) is a kind of RL algorithm and generative model that can generate desirable trajectories (solutions) using the concept of flow networks~\citep{bengio2021flow, bengio2023gflownet}. In GFlowNet, there is a proportional relationship between reward and flow, allowing it to be trained with rewards. A reward is used to train the flow using specific loss function~\citep{bengio2021flow, malkin2022trajectory, madan2023learning} to match flow. GFlowNet generates sequences of actions based on the actions defined in the environment, thereby reaching the terminal state. Through this process, it can generate various solutions with high rewards. The author said that GFlowNet is similar to the human recognition process in terms of stacking thoughts. Solving the ARC task needs human-like AI, thus, the concept of GFlowNet could be a suitable approach for addressing ARC.

\paragraph{Expected usage}

\begin{figure}[htbp!]
\centering
\includegraphics[width=0.8\columnwidth]{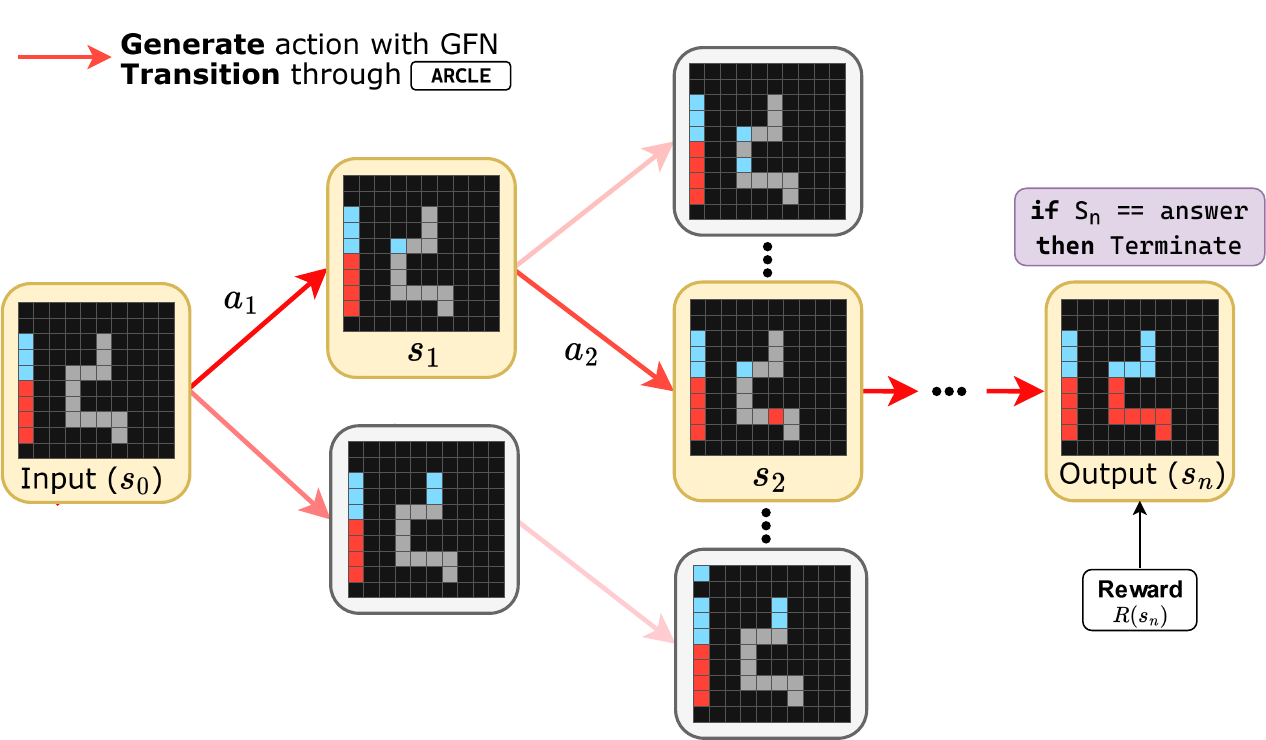}
\caption{A simple architecture of GFlowNet interacting with ARCLE to find paths (solutions or modes). At each iteration, the GFN generates a succession of actions up to the terminal state, exploring the solution. The color saturation of the arrows indicates high and low flow (deeper color indicates higher flow). In our experiments, we were able to guarantee the DAG by coloring the actions one pixel at a time, and we colored the most desirable solution yellow to show that the answer can be found this way.}

\label{fig:GFN_Example}
\end{figure}

ARCLE operates within a discrete action space. Concurrently, GFlowNet, by learning the distribution of actions via a policy network, facilitates the generation of actions while transferring the responsibility for generating the next states to ARCLE. This approach enables a transition-based training methodology for generating action sequences. A crucial element of training GFlowNet is the establishment of a Directed Acyclic Graph (DAG) structure, essential for significantly reducing the search space and enhancing learning efficiency. 
Applying ARCLE to GFlowNet in its basic form could lead to the creation of cycles, thus failing to guarantee DAG structure. A practical method for ensuring a DAG structure involves sequentially coloring one pixel at a time, achievable through a specialized subclass of ARCLE focused on such actions. Considering a $H \times W$ grid size, this leads to a theoretical maximum search space of $10^{H \times W}$. Given GFlowNet's proven efficacy in navigating extensive search spaces in graph combinatorial optimization challenges,~\citep{zhang2023let}, it is expected to outperform other competing algorithms.

The proposed approach aims to construct a DAG to investigate potential solutions, though this methodology mirrors human cognitive processes remains uncertain. Furthermore, even if this method could guarantee DAG structure, there remains a vast search space of $10^{n \times n}$. Since this strategy yields only one solution, it is harder to find an optimal path in contrast to other applications of GFlowNet where solutions encompass a spectrum of possibilities across various paths~\citep{jain2022biological, jain2023multi, zhang2023robust, zhang2023let}. Therefore, for GFlowNet's effective application in solving ARCLE challenges, it is imperative to not only fully utilize ARCLE's actions but also to devise a strategy for DAG construction. Finding a proper way to build a DAG structure could reduce search space significantly and guarantee training effectively.

\newpage
\subsubsection{World Model}
\label{subsubsec:worldmodel}

\paragraph{World Model}

\begin{figure}[h]
\centering
    \begin{subfigure}[t]{.56\linewidth}
        \centering
        \includegraphics[width=\linewidth]{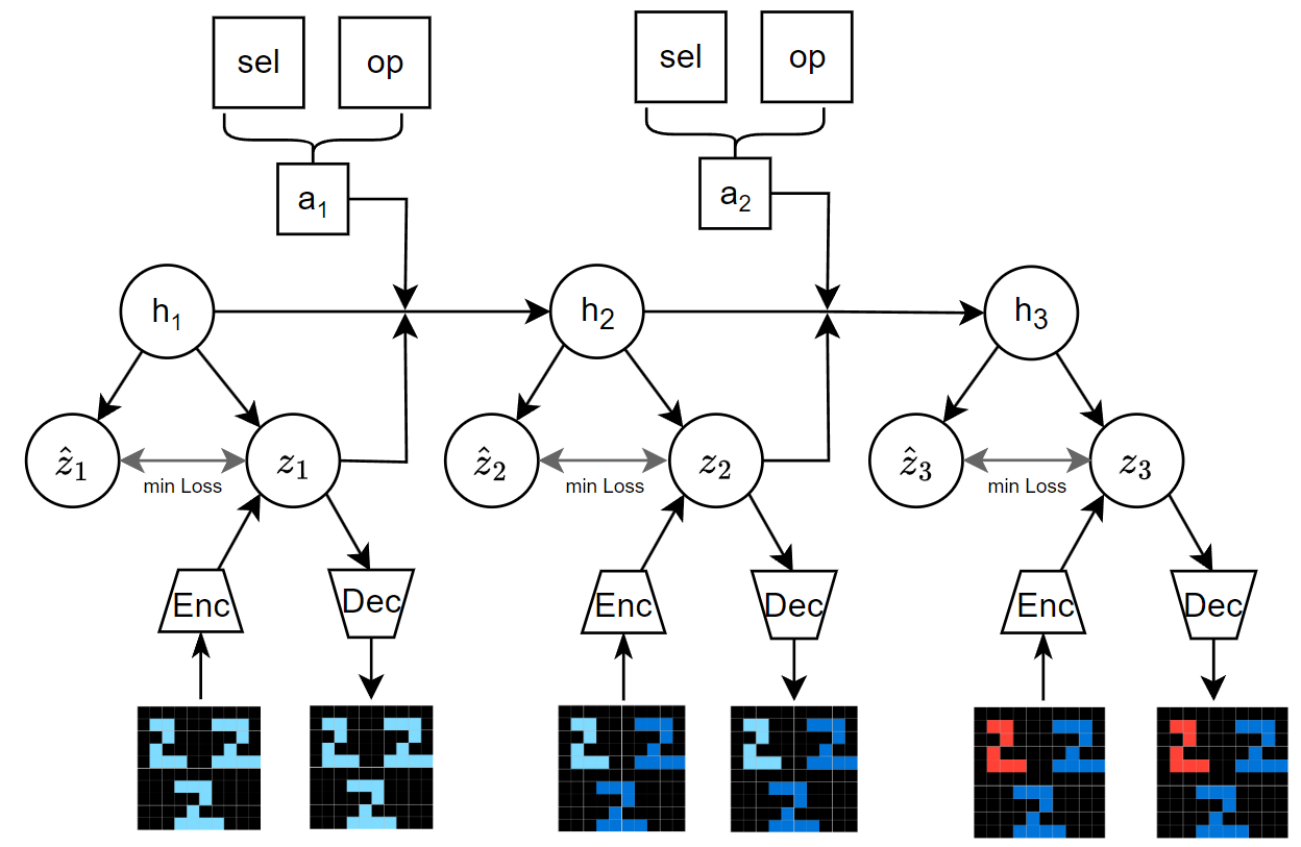}
        \caption{A World Model Architecture}
    \end{subfigure}
    \begin{subfigure}[t]{.42\linewidth}
        \centering
        \includegraphics[width=\linewidth]{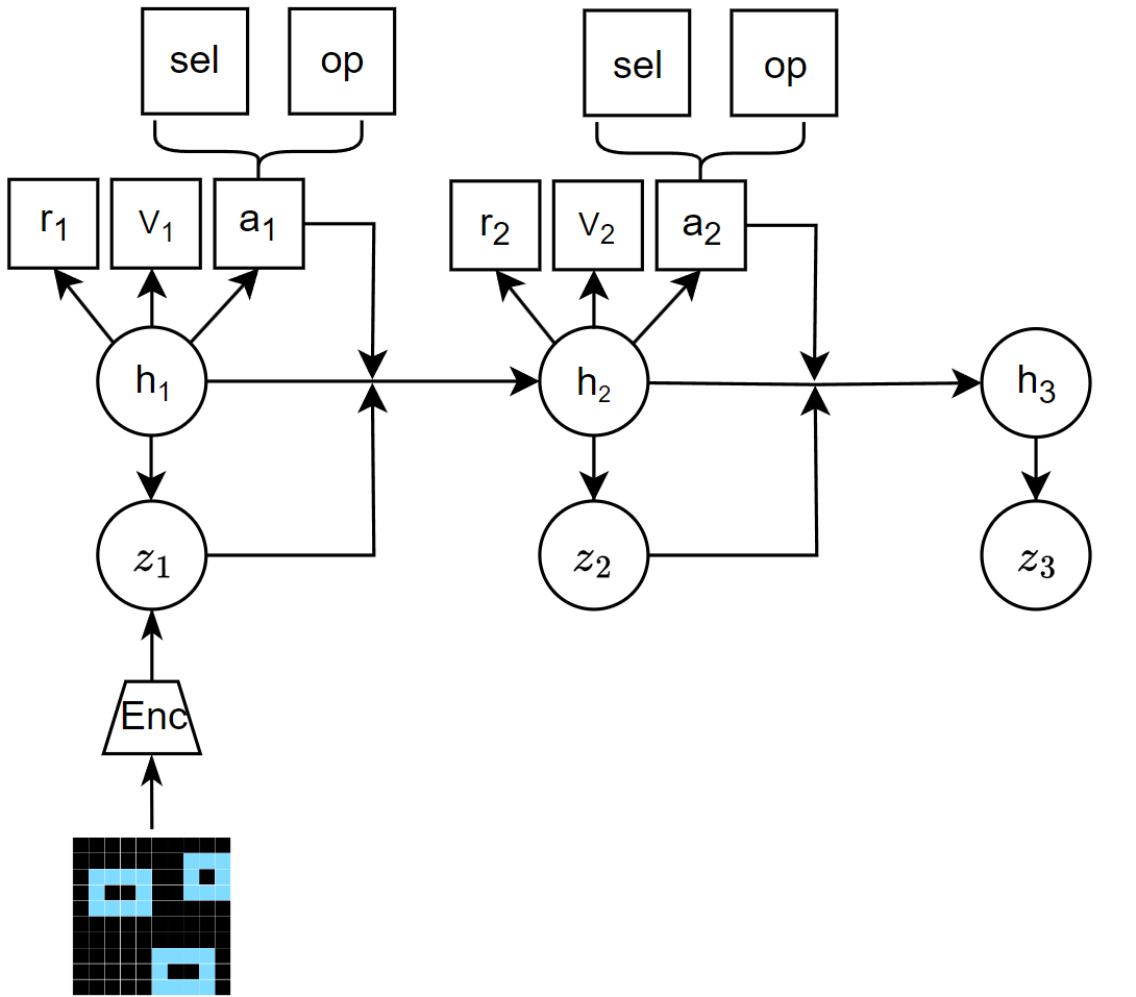}
        \caption{An Actor-Critic Architecture}
    \end{subfigure}
    \caption{An architecture of DreamerV3~\citep{hafner2023mastering}, an advanced version of World Model, solving ARC.}
    \label{fig:form}
\end{figure}

World Model~\citep{ha2018world} is a model-based reinforcement learning algorithm designed to predict models in complex domains where it's difficult to create state spaces and transition functions. To achieve this, the World Model consists of three parts: vision, memory, and controller. Vision is responsible for generating a latent state from given visual images. For instance, in games like Minecraft, where 3D spatial information, items, and various in-game statuses are included in the image, vision transforms it into a vector representing the current state. Memory stores information to predict the next state using the latent state provided by vision. It's known to store prior knowledge inherent to the environment, such as gravity or friction, in vector form. After the learning process, vision and memory function as a kind of model that predicts states and transitions. The controller part is responsible for generating appropriate actions using the given model.

\begin{equation}
\label{eq:world_model_dynamics_update}
\phi_{t+1} = \phi_{t} - \alpha \nabla_{\phi_{t}}\mathcal{L}_{dynamics}(p_{\phi_t}(z_t|x_t)), q_{\phi_t}(z_t|h_t))
\end{equation}

In the above equation, $p_{\phi_t}(z_t|x_t)$ represents vision function which makes latent state $z_t$ given input image $x_t$, whereas $q_{\phi_t}(z_t|h_t)$ represents memory function which makes latent state $z_t$ given transition prediction $h_t$. Subsequently, updated vision and memory are utilized to train the controller using the latent state $z_t$ and transition prediction $h_t$ provided by each, as follows.

\begin{equation}
\label{eq:world_model_controller_update}
\theta_{t+1} = \theta_{t} - \beta \nabla_{\theta_{t}}\mathcal{L}_{controller}(\pi_{\theta_t})
\end{equation}

\paragraph{Expected usage}

The lack of knowledge about which information to extract from input and output grids, as well as the necessary prior knowledge for solving specific types of tasks, is a significant challenge in solving ARC. The Vision part of the World Model excels at extracting important information from visual inputs, while the Memory part is strong at extracting prior knowledge. Therefore, in future ARC research, the architecture of the World Model is expected to play a crucial role. To facilitate future research efforts, this study provides a simple implementation of DreamerV3\citep{hafner2023mastering}, a version of the World Model, applied to ARC.

Furthermore, existing research on World Models has predominantly focused on scenarios where a single agent performs a single task. However, to address the ARC problem, which involves a diverse array of tasks, it might be necessary to consider approaches that incorporate pretraining techniques or apply meta-learning methods. These strategies could potentially enable an agent to adapt to and perform multiple tasks by leveraging prior knowledge or learning how to learn across different tasks.

\end{document}